\pdfoutput=1

\documentclass[11pt]{article}

\usepackage[final]{acl}

\usepackage{times}
\usepackage{latexsym}

\usepackage[T1]{fontenc}

\usepackage{enumitem}
\usepackage{amsmath}

\usepackage[utf8]{inputenc}

\usepackage{microtype}

\usepackage{inconsolata}

\usepackage{graphicx}

\title{Question-Instructed Visual Descriptions \\
for Zero-Shot Video Question Answering}

 \author{David Romero \and Thamar Solorio \\
         MBZUAI  \\ \texttt{\{david.mogrovejo,thamar.solorio\}@mbzuai.ac.ae}}

\begin{document}
\maketitle
\begin{abstract}
We present Q-ViD, a simple approach for video question answering (video QA), that unlike prior methods, which are based on complex architectures, computationally expensive pipelines or use closed models like GPTs, Q-ViD relies on a single instruction-aware open vision-language model (InstructBLIP) to tackle video QA using frame descriptions. Specifically, we create captioning instruction prompts that rely on the target questions about the videos and leverage InstructBLIP to obtain video frame captions that are useful to the task at hand. Subsequently, we form descriptions of the whole video using the question-dependent frame captions, and feed that information, along with a question-answering prompt, to a large language model (LLM). The LLM  is our reasoning module, and performs the final step of multiple-choice QA. Our simple Q-ViD framework achieves competitive or even higher performances than current state of the art models on a diverse range of video QA benchmarks, including NExT-QA, STAR, How2QA, TVQA and IntentQA. Our code is publicly available at: \url{ https://github.com/Daromog/Q-ViD}
\end{abstract}

\section{Introduction}

Recently, vision-language models have shown remarkable performances in image question-answering tasks \citep{Goyal_2017_CVPR, Marino_2019_CVPR, schwenk2022aokvqa}, with models such as Flamingo \citep{NEURIPS2022_960a172b}, BLIP-2 \citep{li2023blip2}, InstructBlip \citep{dai2023instructblip} and mPLUG-Owl \citep{ye2023mplugowl} showing strong reasoning capabilities in the vision-language space. Image captioning \citep{Vinyals_2015_CVPR, Ghandi_2023} is one of the tasks in which these models truly excel, as they can generate detailed linguistic descriptions from images. Different works have leveraged this capability in many ways for zero-shot image-question answering, such as giving linguistic context to images \citep{hu2022promptcap, ghosal-etal-2023-language}, addressing underspecification problems in questions \citep{prasad2023rephrase}, coordination of multiple image captions to complement information \citep{chen2023language}, or by combining captions with other type of linguistic information from the image \citep{berrios2023language}. In this manner, the reasoning capabilities of LLMs can be directly used to reason about the linguistic image descriptions and generate an answer for the given visual question.

This approach has been successful for images, but in the case of video-question answering tasks \citep{lei2018tvqa,li2020hero, xiao2021next, wu2021star, Li_2023_ICCV} this is more challenging. Video possesses multiple image frames that have relationships between each other and involve the recognition of objects, actions, as well as the inference about semantic, temporal, causal reasoning and much more \citep{zhong-etal-2022-video}. Thus, some works \citep{chen2023video, wang2023chatvideo} have focused on using ChatGPT to either ask visual questions to image-language models like BLIP-2 or to respond and retrieve useful information from large datasets with detailed linguistic information from the video. Similarly, \citet{zhang2023simple} have leveraged the reasoning capabilities of GPT-3.5 to create textual summaries from the video, and later perform video QA using only textual information. While others \citep{wang2022language,zeng2022socratic} combine linguistic information from multiple sources such as captions, visual tokenization or even subtitles of input speech. In summary, current methods for video QA rely on any combination of closed LLMs, expensive training regimes, and complex architectures with multiple modules \citep{yang2022frozenbilm, ko2023large, yu2023self, Momeni_2023_ICCV, li2023discovering, zhang2023simple}. In contrast, we introduce Q-ViD a simple \textbf{Q}uestion-Instructed \textbf{Vi}sual \textbf{D}escriptions for video QA approach that relies on an instruction-aware vision-language model, InstructBLIP \citep{dai2023instructblip},  to automatically generate rich specific captions from video frames. In this manner, we effectively turn the video QA task into a text QA task. More specifically, given an input video \(V\) we sample \(n\) number of frames, then, we generate question-specific instructions to prompt the multimodal instruction tuned model to generate captions for each frame. Afterwards, we form a video description by concatenating all the generated question-dependent captions from InstructBLIP, and use it along with the question, options and a question-answering instruction prompt as input to the LLM-based reasoning module that generates an answer to the multiple-choice question about the video. We demonstrate the effectiveness of Q-ViD on five challenging multiple choice video question answering tasks (NExT-QA, STAR, How2QA, TVQA, IntentQA), showing that this simple framework can achieve strong performances comparable with more complex pipelines. Our contributions are summarized as follows:
\begin{itemize}
    \item We propose Q-ViD, a simple gradient-free approach for zero-shot video QA that relies on an open instruction-tuned multimodal model to extract question-specific descriptions of frames to transform the video QA task into a text QA one.
    \item Our approach achieves strong zero-shot performance that is competitive or even superior to more complex architectures such as SeViLa, Internvideo, and Flamingo. It even compares favorably with recent solutions that include GPT APIs, like LloVi and ViperGPT.
\end{itemize}

\begin{figure*}[hbt!]
    \centering
    \includegraphics[width=\linewidth]{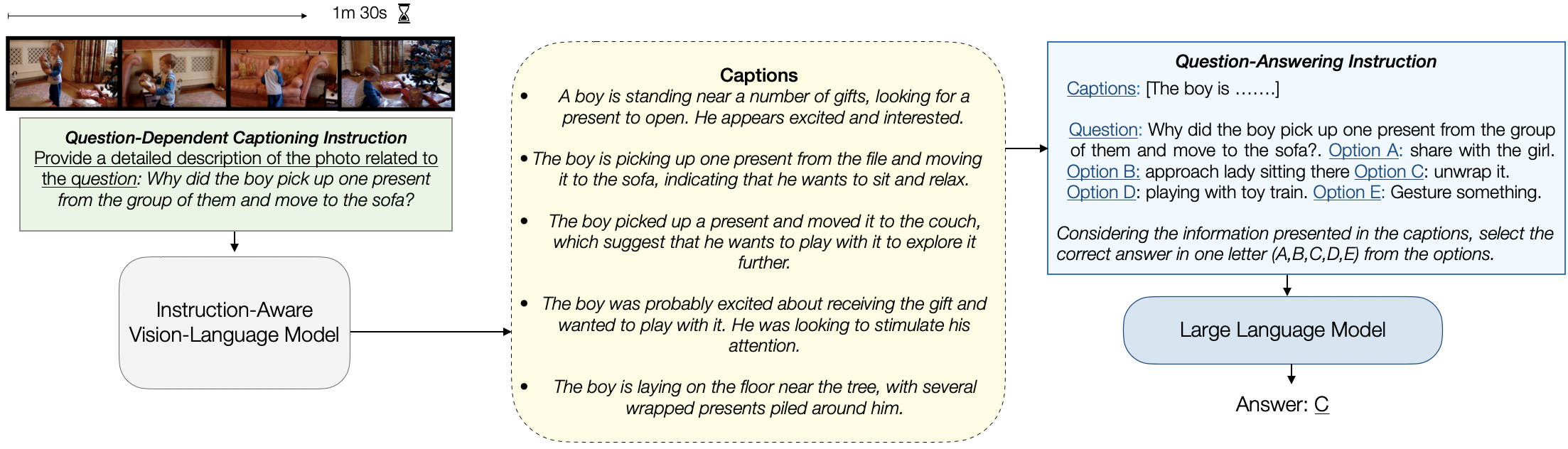}
    \caption{\textbf{Overview of Q-ViD}. We propose relying on a instructed-tuned multimodal model to generate question-dependent frame captions to perform video QA using text. This simple approach achieves competitive results with more complex architectures or GPT-based methods.}
    \label{fig:model}
\end{figure*}

\section{Related Work}

\subsection{Multimodal Pretraining for Video QA}
The strong reasoning capabilities of LLMs \citep{chung2022scaling, touvron2023llama, NEURIPS2020_1457c0d6, hoffmann2022training} in natural language processing tasks has motivated to apply these models for visual understanding. Currently, LLMs have been successfully adapted to understand images \citep{li2023blip2,ye2023mplugowl,chen2023pali}, but applying the same principles for video is more challenging. Approaches for VideoQA rely on image-language models, and adapt those to process video by using fixed amounts of video frames as input \citep{NEURIPS2022_960a172b,yu2023self, yang2022frozenbilm}, or by selecting key-frames from the initial sequence \citep{yu2023self, li2023discovering}. 

Commonly, these works use frozen visual and language models and focus only on modality alignment. Models like Flamingo \citep{NEURIPS2022_960a172b} uses a fixed amount of video frames as input and bridges modalities by training a perceiver resampler and gated attention layers in the Chinchilla LLM \citep{hoffmann2022training}. While others, like SeViLa \citep{yu2023self} relies on BLIP-2 \citep{li2023blip2} for modality alignment, using an intermediate pretrained module called Q-former. SeViLa, first perform key-frame localization and then video QA with Flan-T5 LLMs \citep{chung2022scaling}. On the other hand, other works apart from using frozen vision models, adapt the LLM to visual inputs using adapter tokens \citep{zhang2023llamaadapter} or intermediate trainable modules \citep{houlsby2019parameterefficient}. Models like Flipped-VQA \citep{ko2023large} focuses on adapting LLaMa \citep{touvron2023llama} to video QA by using adapter tokens along with different training objectives to leverage the temporal and causal reasoning abilities of LLMs. Similarly, FrozenBilm \citep{yang2022frozenbilm} exploit the strong zero-shot performance of BILM, a frozen bidirectional language model that is adapted to video QA by using lightweight trainable modules. Despite the success of all these models, they require complex architectures and training regimes, unlike these works we build a simple, gradient-free, approach for zero-shot video QA.

\subsection{Image Captions for Video Understanding}
One of the core strengths of image-language models \citep{NEURIPS2022_960a172b,li2023blip2,dai2023instructblip} is the generation of image captions, thus due to the current strong zero-shot capabilities of LLMs, captions can be directly use to reason about visual content. This has been successfully leveraged in the image-language space for image question-answering with approaches such as Lens \citep{berrios2023language}, Img2LLM \citep{Guo_2023_CVPR} and PromptCat \citep{hu2022promptcap} that gather image captions and other type of linguistic information to answer a visual question. While similar approaches have been taken for videos, the use of large models like GPTs is very common, with models such as ChatCaptioner \citep{chen2023video}, ViperGPT \citep{surís2023vipergpt} , ChatVideo \citep{wang2023chatvideo}, VidIL \citep{wang2022language}, Socratic Models \citep{zeng2022socratic}, and LLoVi \citep{zhang2023simple} 
have been applied for video-language tasks, common methods use GPTs to either interact with image-language models to get visual descriptions, or to make summaries from captions and other type of information such as visual tokenization, subtitles of speech and more. Unlike these approaches, we do not use GPTs or multiple computationally expensive modules in any part of our pipeline to achieve strong zero-shot performance on video QA.

\section{Method}

Recently, vision-language models trained with instruction tuning \citep{dai2023instructblip, zhu2023minigpt, liu2023improved} have shown impressive capabilities to faithfully follow instructions and extract visual representations adapted to the task at hand. Thus, with Q-ViD (Figure ~\ref{fig:model}),we propose to leverage these capabilities for multiple-choice video QA, and turn this task into textual QA using InstructBLIP \citep{dai2023instructblip}. We use a question-dependent captioning prompt as the input instruction, to guide InstructBLIP to generate video frame descriptions that are more relevant for the given question. Afterwards, we reuse the LLM from InstructBLIP and use it as our reasoning module. This LLM (Flan-T5) takes a question-answering prompt as input, that consists of a video description formed by the concatenation of all the question-dependent frame captions, the question, options and a task instruction. Considering that Flan-T5 is also originally trained with instructions, we aim to leverage its reasoning capabilities to correctly answer the question given only the text we just described as input.
Our simple approach does not rely on complex pipelines or closed GPT models, which makes it easy, cheaper and straight forward to use for zero-shot video QA. On the other hand, Q-ViD is flexible and model agnostic, which means we can use any multimodal models available. This section presents our approach in detail. First, we introduce some preliminary information on InstructBLIP, which serves as the foundation of our work, and then we provide a detailed overview for all components from our Q-ViD framework.

\begin{figure*}[hbt!]
    \centering
    \includegraphics[width=\linewidth]{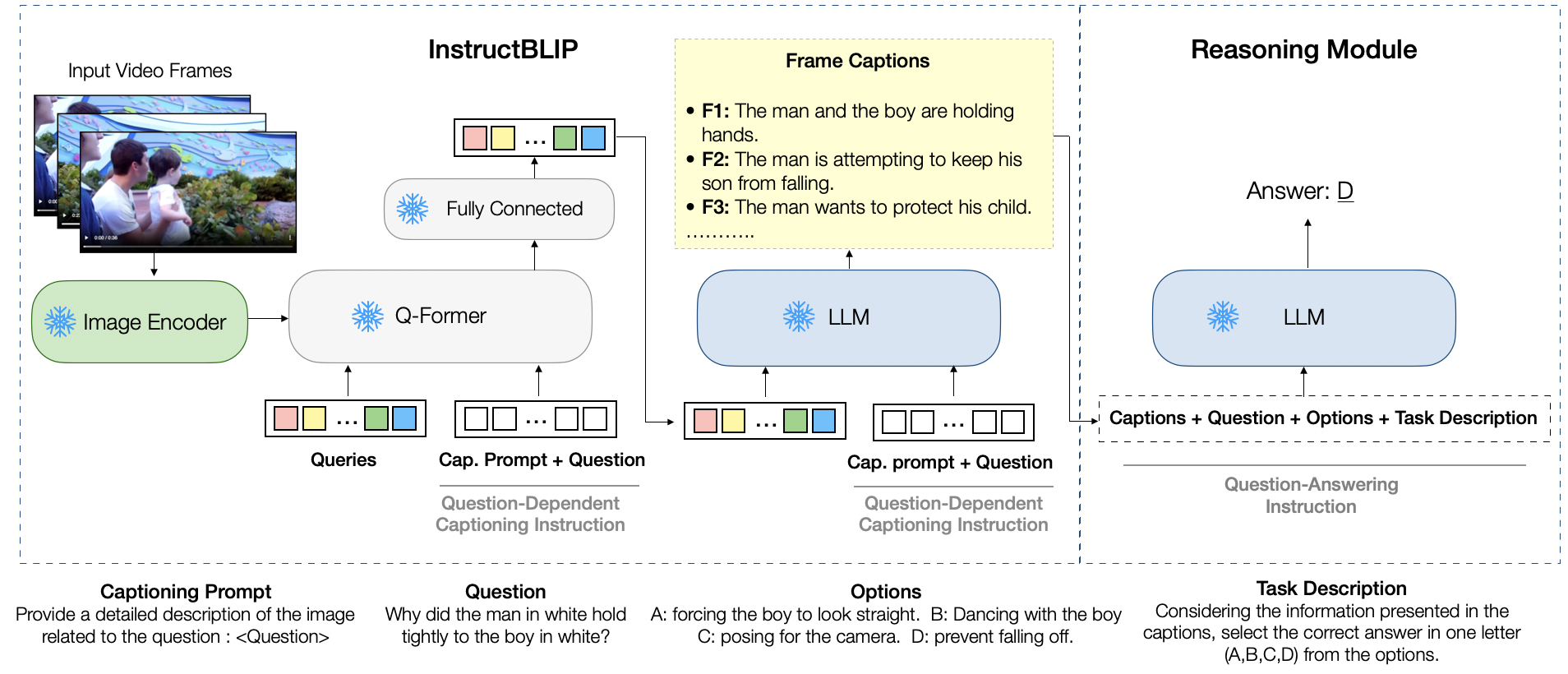}
    \caption{\textbf{Our pipeline for Zero-shot Video QA. } Q-ViD prompts InstructBLIP, to obtain video frame descriptions that are tailored to the question needing answer. }
    \label{fig:architecture}
\end{figure*}

\subsection{Preliminaries: InstructBLIP}
We rely on InstructBLIP \citep{dai2023instructblip} as the foundational architecture of Q-ViD. InstructBLIP is a vision-language instruction tuning framework based on a Query Transformer (Q-former) and frozen vision and language models. Unlike BLIP-2 \citep{li2023blip2},  which is based on an instruction-agnostic approach, InstructBLIP can obtain visual features depending on specific instructions of the task at hand using an instruction-aware Q-former, which in addition to query embeddings, uses instruction tokens to guide the Q-former in extracting specific image features. Subsequently, a LLM (Flan-T5) uses these features to generate visual descriptions depending on the input instructions. In our approach we adapt this model to video, we adopt it to obtain video frame captions that are dependant on the questions of the video QA task, thus, we aim to gather the most important information from each part of the video and use it as input for our reasoning module to answer the given question. Because of our Q-ViD framework is a zero-shot approach, we do not train any part of InstructBLIP, and keep all of its parts frozen.

\subsection{Q-ViD: Generating Frame Descriptions for Video QA}
We focus on automatically generating meaningful captions that can provide enough information about what is happening in the video to the LLM. We assume that if captions for the frames contain relevant information related to the question needing answer, then an LLM should be able to answer the question correctly without additional need for frame/video input. As shown in Figure \ref{fig:architecture}, given an input video \(V\), we  use a uniform sampling strategy and extract a set of \(n\) video frames \(\{f_1,f_2,...,f_n\}\). We then use InstructBLIP, refered as \(I_b\), to obtain instruction-aware visual captions \(c_i\) for each frame \(f_i\), as follows \(c_i=I_b(f_i,E)\), where \(E\) represents the question-dependent captioning instruction. Q-ViD generates \(E\), by concatenating a captioning prompt, referred as \(B\) (e.g \textit{"Provide a detailed description of the image related to the question:"}) and a question, referred as \(Q\) (e.g \textit{"Why did the man in white held tightly to the boy in white?"}), represented as follows \(E=concat(B,Q)\). Specifically, \(E\) is used as input to the Q-former and the LLM of InstructBLIP to obtain specific visual representations and frame descriptions respectively. Thus, we represent the input video \(V\) as a set of question-dependent frame captions \(c=(c_1,c_2,...,c_n)\), where each caption is conformed by a sequence of \(w_m\) words \(c_i=(w_1,w_2,...,w_m)\). In this way, we extract specific textual information from the frames of \(V\), that is going to be useful for the question answering task. Next, we describe the reasoning module of Q-ViD and how these question-dependent captions are used to perform video QA.

\subsection{Q-ViD: Reasoning Module}
We reuse the frozen LLM (Flan-T5) from InstructBLIP and implement it as the reasoning module of Q-ViD. In order to perform video QA using language, we first concatenate the question-dependent frame captions \(C=[c_1,c_2,...,c_n]\) in the same order they appear in the video. Then, we create a question-answering instruction \(L\) as follows: \(L=concat(C,Q,A,T)\). In other words, we concatenate in \(L\) the list of captions \(C\), question \(Q\), possible answers \(A\) and a task description \(T\) (e.g \textit{"Considering the information presented in the captions, select the correct answer in one letter (A,B,C) from the options."}). Our goal is to leverage the LLM reasoning linguistic capabilities by providing a set of captions that were tailored to be relevant for the specific question \(Q\). Our experiments in Section \hyperref[Exp]{\ref*{Exp}}, show that this simple approach works surprisingly well, showing to be competitive, and even superior in some cases, in comparison with more complex pipelines. Next, we describe in more detail the prompts used for question-dependent captioning and video QA. 

\subsection{Q-ViD: Prompt Design} \label{prompt}
First, to get question-dependent captions for each frame, given the question \(Q\) we prompt InstructBLIP with a question-dependent captioning instruction: \textit{\textcolor{darkgray}{"Provide a detailed description of the image related to the question: \{Q\}"}}. This instruction is used along with queries  as input to the frozen Q-Former and LLM modules of InstructBLIP to extract specific visual features and generate question-dependent descriptions. Afterwards, to perform QA with the reasoning module, given the list of captions \(C\) and the list of possible answers \(A=[a_1,...,a_m]\) with \(m\) being the number of options provided in each dataset, we prompt the language model as follows: \textit{\textcolor{darkgray}{"Captions: \{C\} Question: \{Q\}. Option A: a\textsubscript{1}.  Option B: a\textsubscript{2}.  Option C: a\textsubscript{3}. Considering the information presented in the captions, select the correct answer in one letter from the options (A,B,C)"}}. In this prompt, in addition to the list of captions, the question and the list of possible answers, we added a small instruction at the end to specify in detail that a single letter is needed as output. 

\section{Experiments}\label{Exp}
In this section, we present our experiments for zero-shot video QA. First, we describe the datasets we used and the implementation details. Then, we evaluate our approach, compare Q-ViD with other state of the art models for video QA and provide a comprehensive analysis of the model's performance. Lastly, we conduct some ablation studies of Q-ViD regarding the instructions prompt design.

\begin{table*}[h!]
\centering
\small
\begin{tabular}{l@{\hspace{6pt}}c@{\hspace{6pt}}c@{\hspace{6pt}}c@{\hspace{6pt}}c@{\hspace{6pt}}c@{\hspace{6pt}}c@{\hspace{6pt}}c@{\hspace{6pt}}c@{\hspace{6pt}}c@{\hspace{6pt}}cc@{\hspace{6pt}}c}
\hline
\textbf{Models} & \multicolumn{4}{c}{\textbf{NExT-QA}} & \multicolumn{6}{c}{\textbf{STAR}} & \textbf{How2QA} &\textbf{TVQA} \\ \cline{2-5} \cline{7-11} & Tem. & Cau. & Des. & Avg. & & Int. & Seq. & Pre. & Fea. & Avg. \\ \hline 
\textit{GPT-Based Models} \\ 
\textcolor{gray}{ViperGPT} \citep{surís2023vipergpt}  &  -  & - & -  & \textcolor{gray}{60.0} & & -   & -   & -   & -   & - & - & -\\
\textcolor{gray}{LLoVi} \citep{zhang2023simple}  &  \textcolor{gray}{61.0}  & \textcolor{gray}{69.5} & \textcolor{gray}{75.6} & \textcolor{gray}{67.7} & & - & - & - & - & - \\ \hline
Flamingo-9B \citep{NEURIPS2022_960a172b}    & -   & -   & -   & -  &  & -   & -   & -   & - & 41.8 & - & - \\ 
Flamingo-80B \citep{NEURIPS2022_960a172b}   & -   & -   & -   & -  &  & -   & -   & -   & - & 39.7 & - & -  \\ 
FrozenBILM \citep{yang2022frozenbilm} & - & - & - & - &  & - & - & - & - & - & 41.9 & 29.7 \\ 
VFC \citep{Momeni_2023_ICCV}  &  51.6  & 45.4 & 64.1 & 51.6 & &- & - & - & - & - & - & -\\
InternVideo \citep{wang2022internvideo} &  48.0  & 43.4 & 65.1 & 49.1 & & 43.8 & 43.2 & 42.3 & 37.4 & 41.6 & 62.2 & 35.9\\
BLIP-2\raisebox{0.6ex}{voting} \citep{yu2023self} & 59.1 & 61.3 & \underline{74.9} & 62.7 & & 41.8 & 39.7 & 40.2 & 39.5 & 40.3 & 69.8 & 35.7\\
BLIP-2\raisebox{0.6ex}{concat} \citep{yu2023self} & 59.7 & 60.8 & 73.8 & 62.4 & & 45.4 & 41.8 & 41.8 & 40.0 & 42.2 & 70.8 & 36.6 \\
SeViLa \citep{yu2023self} &  \underline{61.3}  & 61.5 & \textbf{75.6} & \underline{63.6} & & \underline{48.3} & 45.0 & \underline{44.4} & 40.8 & 44.6 & \textbf{72.3} & 38.2 \\
VideoChat2 \citep{li2024mvbench} &  57.4  & \underline{61.9} & 69.9 & 61.7 & & \textbf{58.4} & \textbf{60.9} & \textbf{55.3} & \textbf{53.1} & \textbf{59.0} & - & \underline{40.6} \\ \hline
    \rule{0pt}{11pt} \textbf{Q-ViD} (Ours) & \textbf{61.6} & \textbf{67.6} & 72.2 & \textbf{66.3} & & 48.2 & \underline{47.2} & 43.9 & \underline{43.4} & \underline{45.7} & \underline{71.4} & \textbf{41.0}  \\  
\hline
\end{tabular}
\caption{\textbf{Zero-shot results on video question answering.} For fair comparison we \textcolor{gray}{gray out} methods that rely on closed GPTs. We \textbf{bold} the best results, and \underline{underline} the second-best results. Q-ViD shows to be competitive and even outperform some more complex frameworks for zero-shot video QA.}
\label{main_table_1}
\end{table*}

\subsection{Datasets}
To test our approach, we conduct experiments on the following multiple-choice video QA benchmarks. To make comparisons with prior work we use the validation set in NExT-QA, STAR, How2QA and TVQA, meanwhile in IntentQA we use the test set. More details are shown below:

\begin{itemize}[noitemsep]
    \item \textbf{NExT-QA} \citep{xiao2021next}: A benchmark focused on Temporal, Causal and Descriptive reasoning type of questions. Contains 5,440 videos and 48K multiple-choice questions in total. We perform our experiments using the validation set that is conformed by 570 videos and 5K multi-choice questions.
    \item \textbf{STAR} \citep{wu2021star}: A benchmark that evaluates situated reasoning in real-world videos, is focused on interaction, sequence, prediction and feasibility type of questions. It contains 22K situation video clips and 60K questions. We perform evaluations on the validation set with 7K multiple-choice questions.
    \item \textbf{HOW2QA} \citep{li2020hero}: A dataset that consists on 44K question-answering pairs for 22 thousand 60-second clips selected from 9035 videos. We perform experiments on the validation set with 2.8K questions.
    \item \textbf{TVQA} \citep{lei2018tvqa}: A large scale video QA dataset based on six popular TV shows. It has 152K multiple-choice questions and 21K video clips. For our zero-shot evaluations we use the validation set with 15K video-question pairs.
    \item \textbf{IntentQA} \citep{Li_2023_ICCV}: A dataset focused on video intent reasoning. It contains 4K videos and 16K multiple-choice question-answer samples. In this case, we use the test set for our zero-shot evaluations which contains 2K video-question answering samples.
    
\end{itemize}

\subsection{Implementation Details}
For Q-ViD we adopt InstructBLIP-Flan-T5\textsubscript{XXL} with 12.1B parameters, as a default vision encoder it uses VIT-g/14 \citep{Fang_2023_CVPR}, and as language model FlanT5\textsubscript{XXL} \citep{chung2022scaling}. We extract 64 frames per video, as in preliminary experiments this number worked well. For frame captioning, we use a maximum number of 30 tokens per description and top-p sampling with \(top_p=0.7\) to get varied captions. Regarding our reasoning module, we reuse and adopt the corresponding Flan-T5 language model from InstructBLIP. In this case we do not use top-p sampling. Our experiments were conducted using 4 NVIDIA A100 (40GB) GPUs using the Language-Vision Intelligence library LAVIS \citep{li2022lavis} and the released code from SeViLa \citep{yu2023self} .  

\subsection{Overall Performance}\label{ovr}
Table \hyperref[main_table_1]{\ref*{main_table_1}} provides a detailed overview on the performance of Q-ViD on the validation set of NExT-QA, STAR, HOW2QA and TVQA. We compare our approach with current state of the art methods such as SeViLa \citep{yu2023self}, FrozenBILM \citep{yang2022frozenbilm} and VideoChat2 \citep{li2024mvbench}, as well as, with GPT-based models like ViperGPT \citep{surís2023vipergpt} and LloVi \citep{zhang2023simple}. The results obtained from our experiments demonstrate the surprisingly competitive nature of Q-ViD, outperforming or being competitive with previous methods with more complex architectural pipelines such as SeViLa, VideoChat2 and \textcolor{gray}{LLoVi}. For fair comparisons, we \textcolor{gray}{gray out} methods that use GPTs.

Specifically, on \textbf{NExT-QA},  Q-ViD outperforms SeViLa by \textbf{2.7\%} of average accuracy, and achieves almost the same state of the art results of \textcolor{gray}{Llovi}, a framework based of GPT-3.5. Notably, Q-ViD is the best-performing model on causal questions, temporal questions, and overall average performance among methods that are not based on GPTs, showing the ability of this approach to perform action reasoning, which is the target of NExT-QA. With \textbf{STAR}, Q-ViD achieves the second best average accuracy behind VideoChat2, outperforming all other methods like SeViLa by \textbf{1.1\%}, BLIP-2\raisebox{0.6ex}{concat} by \textbf{3.5\%}, InternVideo by \textbf{4.1\%} and Flamingo-80B by \textbf{6\%}. Also note that Q-ViD achieves the second best performances on sequence and feasibility type of question of STAR. Lastly, on \textbf{How2QA} we achieve the second best performance behind SeViLa, and achieves the best overall performance for \textbf{TVQA} with an improvement of \textbf{0.4\%} to the previous best-performing method VideoChat2.

On the other hand, in Table \hyperref[main_table_1]{\ref*{main_table_2}} we evaluate our approach on \textbf{IntentQA}, we use the test set of this benchmark in order to compare with prior works. We take the same comparison made from \citep{zhang2023simple}, and divide the table in two categories, Supervised and Zero-shot approaches. Q-ViD continues showing strong results, greatly outperforming all supervised methods and the SeViLa zero-shot performance by \textbf{2.7\%}. Interestingly, Q-ViD almost achieves the best overall performance from the GPT-based method \textcolor{gray}{Llovi}. These results demonstrate that our approach can be used among different video QA tasks and be able to achieve strong zero-shot performances.

\begin{table}[h!]
\centering
\begin{tabular}{lc}
\hline
\textbf{Models} & \textbf{Acc.(\%)} \\ \hline
\textit{Supervised} \\
HQGA \citep{surís2023vipergpt}  &  47.7  \\
VGT \citep{NEURIPS2022_960a172b}    & 51.3   \\ 
BlindGPT \citep{NEURIPS2022_960a172b}    & 51.6   \\ 
CaVIR \citep{NEURIPS2022_960a172b}    & 57.6   \\ 
\hline
\textit{Zero-shot} \\
SeViLA \citep{yu2023self} & \underline{60.9} \\
\textcolor{gray}{LLoVi} \citep{zhang2023simple} & \textcolor{gray}{64.0} \\ \hline
\rule{0pt}{11pt} \textbf{Q-ViD} (Ours) & \textbf{63.6} \\ 
\hline
\end{tabular}
\caption{\textbf{Performance on IntentQA.} Q-ViD shows to outperform supervised approaches, strong zero-shot baselines like SeViLa and obtain almost the same performance from the GPT-based model LLoVi.}
\label{main_table_2}
\end{table}

\section{Ablation Studies}\label{abl}
In this section, we perform some ablation studies related to the instruction prompt selection for Q-ViD. For these experiments, we chose NExT-QA and STAR as our benchmarks, and report results on the validation sets on each dataset. Specifically, we test two model variations, using InstructBLIP-FlanT5\textsubscript{XL} (Q-ViD\textsubscript{XL}) and the one used to report our main results, InstructBLIP-FlanT5\textsubscript{XXL} (Q-ViD\textsubscript{XXL}), we test different prompts to analyze and compare the use of question-dependent and general descriptive captions. Additionally, we also make some ablation experiments for the question-answering instruction prompt that is used by the reasoning module to perform multi-choice QA. We discuss our findings in detail below.

\subsection{Prompt Analysis}
\begin{figure}[t]
    \centering
    \includegraphics[width=\linewidth]{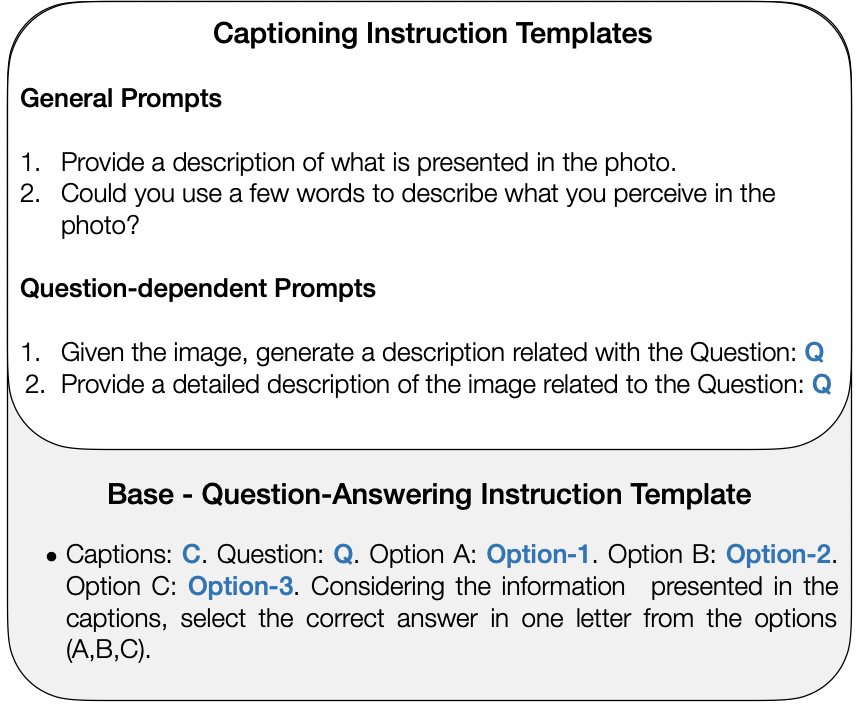}
    \caption{\textbf{Variation of captioning templates.} We focus on comparing general and question-dependent captioning prompts (Top). For both cases we use the same Base QA instruction template (Bottom).}
    \label{fig:baseprompts}
\end{figure}

We focus on analyzing the impact on performance of the Captioning and QA instruction templates in Q-ViD. First, for the captioning instruction template (Figure \ref{fig:baseprompts}), we compare two type of variants:  (1) General prompts and (2) Question-dependent prompts. With general prompts we focus on obtaining general visual descriptions, and with question-dependent prompts on visual information related to the question of the task at hand. In order to test the impact of these captioning prompts, in both cases, we use a Base QA instruction template used as input by the reasoning module (LLM) to perform multiple-choice QA. To leverage as much as possible the instruction-based capabilities of InstructBLIP, we create these prompts based on  similar templates used by this model in its training setup.

\begin{table*}[h!]
\centering
\begin{tabular}{ccccccccccc}
\hline
\textbf{Method} & \multicolumn{4}{c}{\textbf{NExT-QA}} & \multicolumn{6}{c}{\textbf{STAR}} \\ \cline{2-5} \cline{7-11} & Tem. & Cau. & Des. & Avg. & & Int. & Seq. & Pre. & Fea. & Avg. \\ \hline 
\textit{Q-ViD\textsubscript{XL}} \\
(1) General &  \underline{57.3}  & \textbf{60.3} & \textbf{62.0}  & \textbf{60.5} & & \underline{47.0} & \textbf{45.2} & \underline{42.7} & \underline{42.2} & 
\underline{44.3} \\
(2) General & \textbf{57.8} & \underline{60.1} & 60.8 & \underline{60.1} & & \textbf{47.4} & \underline{44.8} & \textbf{44.7} & \textbf{42.8} & \textbf{44.9} \\
(1) Dependent & 55.9 & 59.8 & 57.5 & 59.1 & & 45.0 & 41.7 & 40.5 & 40.2 & 41.8 \\ 
(2) Dependent & 56.6 & 58.8 & \underline{61.1} & 59.0 & & 45.8 & 40.6 & 40.2 & 39.5 & 41.5 \\ \hline

\textit{Q-ViD\textsubscript{XXL}} \\
(1) General &  57.5  & 64.6 & 67.4  & 62.7 & & 44.7 & 39.5 & 42.6 & 36.3 & 40.8 \\
(2) General & 57.1 & 64.8 & 68.0 & 62.8 & & 44.6 & 39.5 & \underline{43.1} & 38.7 & 41.5 \\
(1) Dependent & \textbf{62.0} & \underline{66.5} & \underline{71.2} & \underline{65.8} & & \underline{47.8} & \underline{44.2} & 42.1 & \underline{41.8} & \underline{44.0} \\ 
(2) Dependent & \underline{61.6} & \textbf{67.6} & \textbf{72.2} & \textbf{66.3} & & \textbf{48.2} & \textbf{47.2} & \textbf{43.9} & \textbf{43.4} & \textbf{45.7} \\ \hline

\end{tabular}
\caption{\textbf{Comparing the impact on performance using different Captioning Instruction templates.} We test two variants, General prompts and Question-Dependent prompts. All experiments use the Base QA instruction template. }
\label{cap_prompt}
\end{table*}

Table \hyperref[cap_prompt]{\ref*{cap_prompt}} compares the performance of Q-ViD\textsubscript{XL} and Q-ViD\textsubscript{XXL} using the general, and question-dependent captioning prompts. It can be seen that performance varies between both models. First, Q-ViD\textsubscript{XL} achieves better performances with general captioning prompts, when comparing the best variants of this model, using the (2) General and (1) Dependent prompts, the former further increases the average accuracy by +1.4\% on NExT-QA and +3.1\% on STAR. On the other hand, the same behaviour is not shown using a bigger model, Q-ViD\textsubscript{XXL} achieves significant improvements in average performance by using question-dependent prompts, when comparing its best variants using the (2) General and (2) Dependent prompts, the latter obtains improvements of +3.5\% on NExT-QA and +4.2\% on STAR. Unsurprisingly, Q-ViD\textsubscript{XXL} provides significant performance boosts when compared to its smaller version Q-ViD\textsubscript{XL} achieving better performances on all type of questions in both datasets, showing a better capability to follow instructions, however, \textit{this also demonstrates that using question-dependent prompts to obtain specific information for the task at hand, performs better for zero-shot Video QA than using captioning prompts that obtains general visual descriptions.}

\begin{figure}[hbt!]
    \centering
    \includegraphics[width=\linewidth]{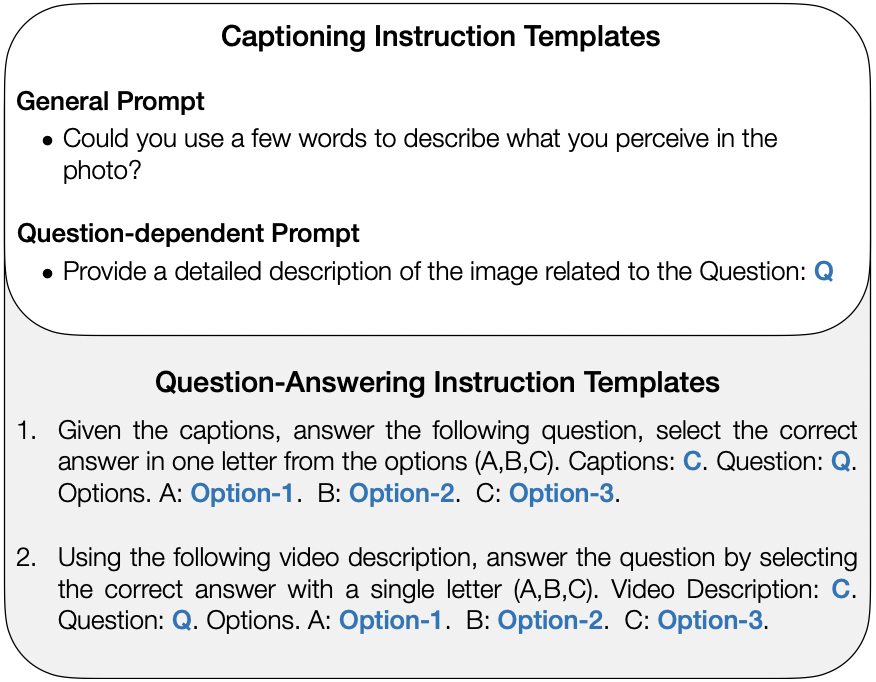}
    \caption{\textbf{Variation of QA prompt templates.} We focus on exploring two more complex and detailed variations for the QA instruction prompt (Bottom). We use the best captioning templates (Top) for Q-ViD\textsubscript{XL} (General) and Q-ViD\textsubscript{XXL} (Dependent).}
    \label{fig:qaprompts}
\end{figure}

Next, in Table \hyperref[qaprompt_perf]{\ref*{qaprompt_perf}} we investigate the impact on performance of the QA Instruction template. We propose two variations that are shown at the bottom of Figure \ref{fig:qaprompts} , in addition to the Base QA template (Figure \ref{fig:baseprompts}). With these new variants we aim to test giving more details to our reasoning module based on Flan-T5, because of this LLM is also a model trained with instructions, we explore if using more complex and detailed QA prompts we can achieve better performances. For this comparison we take the best variants (Table \hyperref[cap_prompt]{\ref*{cap_prompt}}) of Q-ViD\textsubscript{XL} and Q-ViD\textsubscript{XXL} using the (2) General and (2) Dependent captioning prompts respectively for each model, and explore their performances with different QA instruction templates. As shown in Table \hyperref[qaprompt_perf]{\ref*{qaprompt_perf}}, using more complex variants of the initial Base QA Instruction prompt does not have a big impact on performance in any of the models, it even slightly affects the performance in some cases,  showing that the simplest base prompt was enough for the LLM to understand the task. With this ablation study we can highlight the fact that the input instruction used to obtain dedicated frame descriptions is far more important than elaborated question-answering instruction prompts for zero-shot video QA.

\begin{table*}[h!]
\centering
\begin{tabular}{lcccc@{\hspace{6pt}}c@{\hspace{6pt}}c@{\hspace{6pt}}ccc@{\hspace{6pt}}c@{\hspace{6pt}}c@{\hspace{6pt}}c@{\hspace{6pt}}c}
\hline
\textbf{Model} & \multicolumn{2}{c}{\textbf{Templates}} & & \multicolumn{4}{c}{\textbf{NExT-QA}} & & \multicolumn{5}{c}{\textbf{STAR}} \\  \cline{2-3} \cline{5-8} \cline{10-14} &Captioning& QA & & Tem. & Cau. & Des. & Avg. & & Int. & Seq. & Pre. & Fea. & Avg. \\ \hline

 &  & Base & & \textbf{57.8} & 60.1 & \textbf{60.8} & \underline{60.1} & & \underline{47.4} & \underline{44.8} & \textbf{44.7} & \textbf{42.8} & \textbf{44.9} \\
\textit{Q-ViD\textsubscript{XL}} & (2)General & (1)QA & & 56.4  & \underline{60.6} & \underline{58.4}  & \textbf{60.2} & & \textbf{47.7} & \textbf{44.9} & \underline{43.5} & \underline{41.0} & \underline{44.3} \\ 
& & (2)QA & & \underline{56.8}  & \textbf{60.9} & 57.5  & \underline{60.1} & & 47.0 & 44.1 & 43.1 & 40.6 & 43.7 \\ 
\hline

& & Base & & \underline{61.6} & \textbf{67.6} & 72.2 & \textbf{66.3} & & 48.2 & \textbf{47.2} & \textbf{43.9} & \underline{43.4} & \underline{45.7} \\
\textit{Q-ViD\textsubscript{XXL}} & (2)Dependent & (1)QA & & \textbf{61.7}  & \underline{65.8} & \underline{73.7}  & \underline{65.5} & & \underline{48.9} & \underline{46.8} & \underline{43.5} & \textbf{43.8} & \textbf{45.8} \\ 
& & (2)QA & & 61.5  & 65.6 & \textbf{73.9}  & \underline{65.5} & & \textbf{49.1} & 45.9 & 42.9 & 42.6 & 45.1 \\ \hline
\end{tabular}

\caption{\textbf{Performance using different variants for the QA Instruction template.} Base: Refer to the base QA instruction template. For the captioning prompts all models use their best variants, Q-ViD\textsubscript{XL} with (2)General and Q-ViD\textsubscript{XXL} with (2)Dependent. These results suggest that there is no improvements using more complex QA instruction prompts for the reasoning module. }
\label{qaprompt_perf}
\end{table*}

\section{Conclusion}
In this paper, we introduce Q-ViD, a simple, gradient-free approach for zero-shot video QA. Q-ViD turns video QA into textual QA using frame captions. To do so, Q-VID relies on  an instruction-aware visual language model
and uses question-dependent captioning instructions to obtain specific frame descriptions useful for the task at hand.  This information is later used by a reasoning module with a question-answering instruction prompt to perform multiple-choice video QA. Our simple approach achieves competitive or even higher performances than more complex architectures and methods that rely on closed models like the GPT family. In our ablation studies we show that using dedicated instructions to get question-dependent captions works better than common prompts to get general descriptions from frames to perform video QA using captions.

\section*{Limitations}
Even though, Q-ViD has shown to achieve strong performances for zero-shot video question answering, our approach suffers from some limitations. While the adopted instruction-aware multimodal model, InstructBlip, shows to successfully follow instructions from the question and extract meaningful information that can help the reasoning module to come up with the right answer, we have seen that in some cases the model tends to show hallucinations in the captions, or  generate direct short one-word answers instead of a detailed and question-specific description of the image. On the other hand, even though experiments with really long videos are not within the scope of this paper, our approach would no be recommended in those cases, due to the high memory usage that comes with saving detailed frame captions to create an entire video description, which would also affect the LLM-based reasoning module because of the limited amount of tokens allowed as input or due to memory constrains to process the entire video description.

\bibliography{custom}

\begin{thebibliography}{46}
\providecommand{\natexlab}[1]{#1}

\bibitem[{Alayrac et~al.(2022)Alayrac, Donahue, Luc, Miech, Barr, Hasson, Lenc, Mensch, Millican, Reynolds, Ring, Rutherford, Cabi, Han, Gong, Samangooei, Monteiro, Menick, Borgeaud, Brock, Nematzadeh, Sharifzadeh, Bi\'{n}kowski, Barreira, Vinyals, Zisserman, and Simonyan}]{NEURIPS2022_960a172b}
Jean-Baptiste Alayrac, Jeff Donahue, Pauline Luc, Antoine Miech, Iain Barr, Yana Hasson, Karel Lenc, Arthur Mensch, Katherine Millican, Malcolm Reynolds, Roman Ring, Eliza Rutherford, Serkan Cabi, Tengda Han, Zhitao Gong, Sina Samangooei, Marianne Monteiro, Jacob~L Menick, Sebastian Borgeaud, Andy Brock, Aida Nematzadeh, Sahand Sharifzadeh, Miko\l~aj Bi\'{n}kowski, Ricardo Barreira, Oriol Vinyals, Andrew Zisserman, and Kar\'{e}n Simonyan. 2022.
\newblock \href {https://proceedings.neurips.cc/paper_files/paper/2022/file/960a172bc7fbf0177ccccbb411a7d800-Paper-Conference.pdf} {Flamingo: a visual language model for few-shot learning}.
\newblock In \emph{Advances in Neural Information Processing Systems}, volume~35, pages 23716--23736. Curran Associates, Inc.

\bibitem[{Berrios et~al.(2023)Berrios, Mittal, Thrush, Kiela, and Singh}]{berrios2023language}
William Berrios, Gautam Mittal, Tristan Thrush, Douwe Kiela, and Amanpreet Singh. 2023.
\newblock \href {https://arxiv.org/abs/2306.16410} {Towards language models that can see: Computer vision through the lens of natural language}.
\newblock \emph{Preprint}, arXiv:2306.16410.

\bibitem[{Brown et~al.(2020)Brown, Mann, Ryder, Subbiah, Kaplan, Dhariwal, Neelakantan, Shyam, Sastry, Askell, Agarwal, Herbert-Voss, Krueger, Henighan, Child, Ramesh, Ziegler, Wu, Winter, Hesse, Chen, Sigler, Litwin, Gray, Chess, Clark, Berner, McCandlish, Radford, Sutskever, and Amodei}]{NEURIPS2020_1457c0d6}
Tom Brown, Benjamin Mann, Nick Ryder, Melanie Subbiah, Jared~D Kaplan, Prafulla Dhariwal, Arvind Neelakantan, Pranav Shyam, Girish Sastry, Amanda Askell, Sandhini Agarwal, Ariel Herbert-Voss, Gretchen Krueger, Tom Henighan, Rewon Child, Aditya Ramesh, Daniel Ziegler, Jeffrey Wu, Clemens Winter, Chris Hesse, Mark Chen, Eric Sigler, Mateusz Litwin, Scott Gray, Benjamin Chess, Jack Clark, Christopher Berner, Sam McCandlish, Alec Radford, Ilya Sutskever, and Dario Amodei. 2020.
\newblock \href {https://proceedings.neurips.cc/paper_files/paper/2020/file/1457c0d6bfcb4967418bfb8ac142f64a-Paper.pdf} {Language models are few-shot learners}.
\newblock In \emph{Advances in Neural Information Processing Systems}, volume~33, pages 1877--1901. Curran Associates, Inc.

\bibitem[{Chen et~al.(2023{\natexlab{a}})Chen, Zhu, Haydarov, Li, and Elhoseiny}]{chen2023video}
Jun Chen, Deyao Zhu, Kilichbek Haydarov, Xiang Li, and Mohamed Elhoseiny. 2023{\natexlab{a}}.
\newblock Video chatcaptioner: Towards the enriched spatiotemporal descriptions.
\newblock \emph{arXiv preprint arXiv:2304.04227}.

\bibitem[{Chen et~al.(2023{\natexlab{b}})Chen, Li, Shen, Yang, Li, Keutzer, Darrell, and Liu}]{chen2023language}
Liangyu Chen, Bo~Li, Sheng Shen, Jingkang Yang, Chunyuan Li, Kurt Keutzer, Trevor Darrell, and Ziwei Liu. 2023{\natexlab{b}}.
\newblock \href {https://openreview.net/forum?id=kdHpWogtX6Y} {Language models are visual reasoning coordinators}.
\newblock In \emph{ICLR 2023 Workshop on Mathematical and Empirical Understanding of Foundation Models}.

\bibitem[{Chen et~al.(2023{\natexlab{c}})Chen, Wang, Changpinyo, Piergiovanni, Padlewski, Salz, Goodman, Grycner, Mustafa, Beyer, Kolesnikov, Puigcerver, Ding, Rong, Akbari, Mishra, Xue, Thapliyal, Bradbury, Kuo, Seyedhosseini, Jia, Ayan, Riquelme, Steiner, Angelova, Zhai, Houlsby, and Soricut}]{chen2023pali}
Xi~Chen, Xiao Wang, Soravit Changpinyo, AJ~Piergiovanni, Piotr Padlewski, Daniel Salz, Sebastian Goodman, Adam Grycner, Basil Mustafa, Lucas Beyer, Alexander Kolesnikov, Joan Puigcerver, Nan Ding, Keran Rong, Hassan Akbari, Gaurav Mishra, Linting Xue, Ashish Thapliyal, James Bradbury, Weicheng Kuo, Mojtaba Seyedhosseini, Chao Jia, Burcu~Karagol Ayan, Carlos Riquelme, Andreas Steiner, Anelia Angelova, Xiaohua Zhai, Neil Houlsby, and Radu Soricut. 2023{\natexlab{c}}.
\newblock \href {https://arxiv.org/abs/2209.06794} {Pali: A jointly-scaled multilingual language-image model}.
\newblock \emph{Preprint}, arXiv:2209.06794.

\bibitem[{Chung et~al.(2022)Chung, Hou, Longpre, Zoph, Tay, Fedus, Li, Wang, Dehghani, Brahma, Webson, Gu, Dai, Suzgun, Chen, Chowdhery, Castro-Ros, Pellat, Robinson, Valter, Narang, Mishra, Yu, Zhao, Huang, Dai, Yu, Petrov, Chi, Dean, Devlin, Roberts, Zhou, Le, and Wei}]{chung2022scaling}
Hyung~Won Chung, Le~Hou, Shayne Longpre, Barret Zoph, Yi~Tay, William Fedus, Yunxuan Li, Xuezhi Wang, Mostafa Dehghani, Siddhartha Brahma, Albert Webson, Shixiang~Shane Gu, Zhuyun Dai, Mirac Suzgun, Xinyun Chen, Aakanksha Chowdhery, Alex Castro-Ros, Marie Pellat, Kevin Robinson, Dasha Valter, Sharan Narang, Gaurav Mishra, Adams Yu, Vincent Zhao, Yanping Huang, Andrew Dai, Hongkun Yu, Slav Petrov, Ed~H. Chi, Jeff Dean, Jacob Devlin, Adam Roberts, Denny Zhou, Quoc~V. Le, and Jason Wei. 2022.
\newblock \href {https://arxiv.org/abs/2210.11416} {Scaling instruction-finetuned language models}.
\newblock \emph{Preprint}, arXiv:2210.11416.

\bibitem[{Dai et~al.(2023)Dai, Li, Li, Tiong, Zhao, Wang, Li, Fung, and Hoi}]{dai2023instructblip}
Wenliang Dai, Junnan Li, Dongxu Li, Anthony Meng~Huat Tiong, Junqi Zhao, Weisheng Wang, Boyang Li, Pascale Fung, and Steven Hoi. 2023.
\newblock \href {https://arxiv.org/abs/2305.06500} {Instructblip: Towards general-purpose vision-language models with instruction tuning}.
\newblock \emph{Preprint}, arXiv:2305.06500.

\bibitem[{Fang et~al.(2023)Fang, Wang, Xie, Sun, Wu, Wang, Huang, Wang, and Cao}]{Fang_2023_CVPR}
Yuxin Fang, Wen Wang, Binhui Xie, Quan Sun, Ledell Wu, Xinggang Wang, Tiejun Huang, Xinlong Wang, and Yue Cao. 2023.
\newblock Eva: Exploring the limits of masked visual representation learning at scale.
\newblock In \emph{Proceedings of the IEEE/CVF Conference on Computer Vision and Pattern Recognition (CVPR)}, pages 19358--19369.

\bibitem[{Ghandi et~al.(2023)Ghandi, Pourreza, and Mahyar}]{Ghandi_2023}
Taraneh Ghandi, Hamidreza Pourreza, and Hamidreza Mahyar. 2023.
\newblock \href {https://doi.org/10.1145/3617592} {Deep learning approaches on image captioning: A review}.
\newblock \emph{ACM Computing Surveys}, 56(3):1–39.

\bibitem[{Ghosal et~al.(2023)Ghosal, Majumder, Lee, Mihalcea, and Poria}]{ghosal-etal-2023-language}
Deepanway Ghosal, Navonil Majumder, Roy Lee, Rada Mihalcea, and Soujanya Poria. 2023.
\newblock \href {https://doi.org/10.18653/v1/2023.findings-emnlp.809} {Language guided visual question answering: Elevate your multimodal language model using knowledge-enriched prompts}.
\newblock In \emph{Findings of the Association for Computational Linguistics: EMNLP 2023}, pages 12096--12102, Singapore. Association for Computational Linguistics.

\bibitem[{Goyal et~al.(2017)Goyal, Khot, Summers-Stay, Batra, and Parikh}]{Goyal_2017_CVPR}
Yash Goyal, Tejas Khot, Douglas Summers-Stay, Dhruv Batra, and Devi Parikh. 2017.
\newblock Making the v in vqa matter: Elevating the role of image understanding in visual question answering.
\newblock In \emph{Proceedings of the IEEE Conference on Computer Vision and Pattern Recognition (CVPR)}.

\bibitem[{Guo et~al.(2023)Guo, Li, Li, Tiong, Li, Tao, and Hoi}]{Guo_2023_CVPR}
Jiaxian Guo, Junnan Li, Dongxu Li, Anthony Meng~Huat Tiong, Boyang Li, Dacheng Tao, and Steven Hoi. 2023.
\newblock From images to textual prompts: Zero-shot visual question answering with frozen large language models.
\newblock In \emph{Proceedings of the IEEE/CVF Conference on Computer Vision and Pattern Recognition (CVPR)}, pages 10867--10877.

\bibitem[{Hoffmann et~al.(2022)Hoffmann, Borgeaud, Mensch, Buchatskaya, Cai, Rutherford, de~Las~Casas, Hendricks, Welbl, Clark, Hennigan, Noland, Millican, van~den Driessche, Damoc, Guy, Osindero, Simonyan, Elsen, Rae, Vinyals, and Sifre}]{hoffmann2022training}
Jordan Hoffmann, Sebastian Borgeaud, Arthur Mensch, Elena Buchatskaya, Trevor Cai, Eliza Rutherford, Diego de~Las~Casas, Lisa~Anne Hendricks, Johannes Welbl, Aidan Clark, Tom Hennigan, Eric Noland, Katie Millican, George van~den Driessche, Bogdan Damoc, Aurelia Guy, Simon Osindero, Karen Simonyan, Erich Elsen, Jack~W. Rae, Oriol Vinyals, and Laurent Sifre. 2022.
\newblock \href {https://arxiv.org/abs/2203.15556} {Training compute-optimal large language models}.
\newblock \emph{Preprint}, arXiv:2203.15556.

\bibitem[{Houlsby et~al.(2019)Houlsby, Giurgiu, Jastrzebski, Morrone, de~Laroussilhe, Gesmundo, Attariyan, and Gelly}]{houlsby2019parameterefficient}
Neil Houlsby, Andrei Giurgiu, Stanislaw Jastrzebski, Bruna Morrone, Quentin de~Laroussilhe, Andrea Gesmundo, Mona Attariyan, and Sylvain Gelly. 2019.
\newblock \href {https://arxiv.org/abs/1902.00751} {Parameter-efficient transfer learning for nlp}.
\newblock \emph{Preprint}, arXiv:1902.00751.

\bibitem[{Hu et~al.(2022)Hu, Hua, Yang, Shi, Smith, and Luo}]{hu2022promptcap}
Yushi* Hu, Hang* Hua, Zhengyuan Yang, Weijia Shi, Noah~A Smith, and Jiebo Luo. 2022.
\newblock Promptcap: Prompt-guided task-aware image captioning.
\newblock \emph{arXiv preprint arXiv:2211.09699}.

\bibitem[{Ko et~al.(2023)Ko, Lee, Kang, Roh, and Kim}]{ko2023large}
Dohwan Ko, Ji~Soo Lee, Wooyoung Kang, Byungseok Roh, and Hyunwoo~J Kim. 2023.
\newblock Large language models are temporal and causal reasoners for video question answering.
\newblock In \emph{EMNLP}.

\bibitem[{Lei et~al.(2018)Lei, Yu, Bansal, and Berg}]{lei2018tvqa}
Jie Lei, Licheng Yu, Mohit Bansal, and Tamara~L Berg. 2018.
\newblock Tvqa: Localized, compositional video question answering.
\newblock In \emph{EMNLP}.

\bibitem[{Li et~al.(2022)Li, Li, Le, Wang, Savarese, and Hoi}]{li2022lavis}
Dongxu Li, Junnan Li, Hung Le, Guangsen Wang, Silvio Savarese, and Steven C.~H. Hoi. 2022.
\newblock \href {https://arxiv.org/abs/2209.09019} {Lavis: A library for language-vision intelligence}.
\newblock \emph{Preprint}, arXiv:2209.09019.

\bibitem[{Li et~al.(2023{\natexlab{a}})Li, Wei, Han, and Fan}]{Li_2023_ICCV}
Jiapeng Li, Ping Wei, Wenjuan Han, and Lifeng Fan. 2023{\natexlab{a}}.
\newblock Intentqa: Context-aware video intent reasoning.
\newblock In \emph{Proceedings of the IEEE/CVF International Conference on Computer Vision (ICCV)}, pages 11963--11974.

\bibitem[{Li et~al.(2023{\natexlab{b}})Li, Li, Savarese, and Hoi}]{li2023blip2}
Junnan Li, Dongxu Li, Silvio Savarese, and Steven Hoi. 2023{\natexlab{b}}.
\newblock \href {https://arxiv.org/abs/2301.12597} {Blip-2: Bootstrapping language-image pre-training with frozen image encoders and large language models}.
\newblock \emph{Preprint}, arXiv:2301.12597.

\bibitem[{Li et~al.(2024)Li, Wang, He, Li, Wang, Liu, Wang, Xu, Chen, Luo, Wang, and Qiao}]{li2024mvbench}
Kunchang Li, Yali Wang, Yinan He, Yizhuo Li, Yi~Wang, Yi~Liu, Zun Wang, Jilan Xu, Guo Chen, Ping Luo, Limin Wang, and Yu~Qiao. 2024.
\newblock \href {https://arxiv.org/abs/2311.17005} {Mvbench: A comprehensive multi-modal video understanding benchmark}.
\newblock \emph{Preprint}, arXiv:2311.17005.

\bibitem[{Li et~al.(2020)Li, Chen, Cheng, Gan, Yu, and Liu}]{li2020hero}
Linjie Li, Yen-Chun Chen, Yu~Cheng, Zhe Gan, Licheng Yu, and Jingjing Liu. 2020.
\newblock Hero: Hierarchical encoder for video+ language omni-representation pre-training.
\newblock In \emph{EMNLP}.

\bibitem[{Li et~al.(2023{\natexlab{c}})Li, Xiao, Feng, Wang, and Chua}]{li2023discovering}
Yicong Li, Junbin Xiao, Chun Feng, Xiang Wang, and Tat-Seng Chua. 2023{\natexlab{c}}.
\newblock \href {https://arxiv.org/abs/2307.12058} {Discovering spatio-temporal rationales for video question answering}.
\newblock \emph{Preprint}, arXiv:2307.12058.

\bibitem[{Liu et~al.(2023)Liu, Li, Li, and Lee}]{liu2023improved}
Haotian Liu, Chunyuan Li, Yuheng Li, and Yong~Jae Lee. 2023.
\newblock \href {https://arxiv.org/abs/2310.03744} {Improved baselines with visual instruction tuning}.
\newblock \emph{Preprint}, arXiv:2310.03744.

\bibitem[{Marino et~al.(2019)Marino, Rastegari, Farhadi, and Mottaghi}]{Marino_2019_CVPR}
Kenneth Marino, Mohammad Rastegari, Ali Farhadi, and Roozbeh Mottaghi. 2019.
\newblock Ok-vqa: A visual question answering benchmark requiring external knowledge.
\newblock In \emph{Proceedings of the IEEE/CVF Conference on Computer Vision and Pattern Recognition (CVPR)}.

\bibitem[{Momeni et~al.(2023)Momeni, Caron, Nagrani, Zisserman, and Schmid}]{Momeni_2023_ICCV}
Liliane Momeni, Mathilde Caron, Arsha Nagrani, Andrew Zisserman, and Cordelia Schmid. 2023.
\newblock Verbs in action: Improving verb understanding in video-language models.
\newblock In \emph{Proceedings of the IEEE/CVF International Conference on Computer Vision (ICCV)}, pages 15579--15591.

\bibitem[{Paszke et~al.(2019)Paszke, Gross, Massa, Lerer, Bradbury, Chanan, Killeen, Lin, Gimelshein, Antiga, Desmaison, Köpf, Yang, DeVito, Raison, Tejani, Chilamkurthy, Steiner, Fang, Bai, and Chintala}]{paszke2019pytorch}
Adam Paszke, Sam Gross, Francisco Massa, Adam Lerer, James Bradbury, Gregory Chanan, Trevor Killeen, Zeming Lin, Natalia Gimelshein, Luca Antiga, Alban Desmaison, Andreas Köpf, Edward Yang, Zach DeVito, Martin Raison, Alykhan Tejani, Sasank Chilamkurthy, Benoit Steiner, Lu~Fang, Junjie Bai, and Soumith Chintala. 2019.
\newblock \href {https://arxiv.org/abs/1912.01703} {Pytorch: An imperative style, high-performance deep learning library}.
\newblock \emph{Preprint}, arXiv:1912.01703.

\bibitem[{Prasad et~al.(2023)Prasad, Stengel-Eskin, and Bansal}]{prasad2023rephrase}
Archiki Prasad, Elias Stengel-Eskin, and Mohit Bansal. 2023.
\newblock \href {https://arxiv.org/abs/2310.05861} {Rephrase, augment, reason: Visual grounding of questions for vision-language models}.
\newblock \emph{Preprint}, arXiv:2310.05861.

\bibitem[{Schwenk et~al.(2022)Schwenk, Khandelwal, Clark, Marino, and Mottaghi}]{schwenk2022aokvqa}
Dustin Schwenk, Apoorv Khandelwal, Christopher Clark, Kenneth Marino, and Roozbeh Mottaghi. 2022.
\newblock \href {https://arxiv.org/abs/2206.01718} {A-okvqa: A benchmark for visual question answering using world knowledge}.
\newblock \emph{Preprint}, arXiv:2206.01718.

\bibitem[{Surís et~al.(2023)Surís, Menon, and Vondrick}]{surís2023vipergpt}
Dídac Surís, Sachit Menon, and Carl Vondrick. 2023.
\newblock \href {https://arxiv.org/abs/2303.08128} {Vipergpt: Visual inference via python execution for reasoning}.
\newblock \emph{Preprint}, arXiv:2303.08128.

\bibitem[{Touvron et~al.(2023)Touvron, Lavril, Izacard, Martinet, Lachaux, Lacroix, Rozière, Goyal, Hambro, Azhar, Rodriguez, Joulin, Grave, and Lample}]{touvron2023llama}
Hugo Touvron, Thibaut Lavril, Gautier Izacard, Xavier Martinet, Marie-Anne Lachaux, Timothée Lacroix, Baptiste Rozière, Naman Goyal, Eric Hambro, Faisal Azhar, Aurelien Rodriguez, Armand Joulin, Edouard Grave, and Guillaume Lample. 2023.
\newblock \href {https://arxiv.org/abs/2302.13971} {Llama: Open and efficient foundation language models}.
\newblock \emph{Preprint}, arXiv:2302.13971.

\bibitem[{Vinyals et~al.(2015)Vinyals, Toshev, Bengio, and Erhan}]{Vinyals_2015_CVPR}
Oriol Vinyals, Alexander Toshev, Samy Bengio, and Dumitru Erhan. 2015.
\newblock Show and tell: A neural image caption generator.
\newblock In \emph{Proceedings of the IEEE Conference on Computer Vision and Pattern Recognition (CVPR)}.

\bibitem[{Wang et~al.(2023)Wang, Chen, Luo, Dai, Yuan, Wu, and Jiang}]{wang2023chatvideo}
Junke Wang, Dongdong Chen, Chong Luo, Xiyang Dai, Lu~Yuan, Zuxuan Wu, and Yu-Gang Jiang. 2023.
\newblock \href {https://arxiv.org/abs/2304.14407} {Chatvideo: A tracklet-centric multimodal and versatile video understanding system}.
\newblock \emph{Preprint}, arXiv:2304.14407.

\bibitem[{Wang et~al.(2022{\natexlab{a}})Wang, Li, Li, He, Huang, Zhao, Zhang, Xu, Liu, Wang, Xing, Chen, Pan, Yu, Wang, Wang, and Qiao}]{wang2022internvideo}
Yi~Wang, Kunchang Li, Yizhuo Li, Yinan He, Bingkun Huang, Zhiyu Zhao, Hongjie Zhang, Jilan Xu, Yi~Liu, Zun Wang, Sen Xing, Guo Chen, Junting Pan, Jiashuo Yu, Yali Wang, Limin Wang, and Yu~Qiao. 2022{\natexlab{a}}.
\newblock \href {https://arxiv.org/abs/2212.03191} {Internvideo: General video foundation models via generative and discriminative learning}.
\newblock \emph{Preprint}, arXiv:2212.03191.

\bibitem[{Wang et~al.(2022{\natexlab{b}})Wang, Li, Xu, Zhou, Lei, Lin, Wang, Yang, Zhu, Hoiem, Chang, Bansal, and Ji}]{wang2022language}
Zhenhailong Wang, Manling Li, Ruochen Xu, Luowei Zhou, Jie Lei, Xudong Lin, Shuohang Wang, Ziyi Yang, Chenguang Zhu, Derek Hoiem, Shih-Fu Chang, Mohit Bansal, and Heng Ji. 2022{\natexlab{b}}.
\newblock \href {https://arxiv.org/abs/2205.10747} {Language models with image descriptors are strong few-shot video-language learners}.
\newblock \emph{Preprint}, arXiv:2205.10747.

\bibitem[{Wu et~al.(2021)Wu, Yu, Chen, Tenenbaum, and Gan}]{wu2021star}
Bo~Wu, Shoubin Yu, Zhenfang Chen, Joshua~B. Tenenbaum, and Chuang Gan. 2021.
\newblock \href {https://openreview.net/forum?id=EfgNF5-ZAjM} {{STAR}: A benchmark for situated reasoning in real-world videos}.
\newblock In \emph{Thirty-fifth Conference on Neural Information Processing Systems Datasets and Benchmarks Track (Round 2)}.

\bibitem[{Xiao et~al.(2021)Xiao, Shang, Yao, and Chua}]{xiao2021next}
Junbin Xiao, Xindi Shang, Angela Yao, and Tat-Seng Chua. 2021.
\newblock Next-qa: Next phase of question-answering to explaining temporal actions.
\newblock In \emph{Proceedings of the IEEE/CVF Conference on Computer Vision and Pattern Recognition (CVPR)}, pages 9777--9786.

\bibitem[{Yang et~al.(2022)Yang, Miech, Sivic, Laptev, and Schmid}]{yang2022frozenbilm}
Antoine Yang, Antoine Miech, Josef Sivic, Ivan Laptev, and Cordelia Schmid. 2022.
\newblock Zero-shot video question answering via frozen bidirectional language models.
\newblock In \emph{NeurIPS}.

\bibitem[{Ye et~al.(2023)Ye, Xu, Xu, Ye, Yan, Zhou, Wang, Hu, Shi, Shi, Jiang, Li, Xu, Chen, Tian, Qi, Zhang, and Huang}]{ye2023mplugowl}
Qinghao Ye, Haiyang Xu, Guohai Xu, Jiabo Ye, Ming Yan, Yiyang Zhou, Junyang Wang, Anwen Hu, Pengcheng Shi, Yaya Shi, Chaoya Jiang, Chenliang Li, Yuanhong Xu, Hehong Chen, Junfeng Tian, Qian Qi, Ji~Zhang, and Fei Huang. 2023.
\newblock \href {https://arxiv.org/abs/2304.14178} {mplug-owl: Modularization empowers large language models with multimodality}.
\newblock \emph{Preprint}, arXiv:2304.14178.

\bibitem[{Yu et~al.(2023)Yu, Cho, Yadav, and Bansal}]{yu2023self}
Shoubin Yu, Jaemin Cho, Prateek Yadav, and Mohit Bansal. 2023.
\newblock Self-chained image-language model for video localization and question answering.
\newblock In \emph{NeurIPS}.

\bibitem[{Zeng et~al.(2022)Zeng, Attarian, Ichter, Choromanski, Wong, Welker, Tombari, Purohit, Ryoo, Sindhwani, Lee, Vanhoucke, and Florence}]{zeng2022socratic}
Andy Zeng, Maria Attarian, Brian Ichter, Krzysztof Choromanski, Adrian Wong, Stefan Welker, Federico Tombari, Aveek Purohit, Michael Ryoo, Vikas Sindhwani, Johnny Lee, Vincent Vanhoucke, and Pete Florence. 2022.
\newblock \href {https://arxiv.org/abs/2204.00598} {Socratic models: Composing zero-shot multimodal reasoning with language}.
\newblock \emph{Preprint}, arXiv:2204.00598.

\bibitem[{Zhang et~al.(2023{\natexlab{a}})Zhang, Lu, Islam, Wang, Yu, Bansal, and Bertasius}]{zhang2023simple}
Ce~Zhang, Taixi Lu, Md~Mohaiminul Islam, Ziyang Wang, Shoubin Yu, Mohit Bansal, and Gedas Bertasius. 2023{\natexlab{a}}.
\newblock \href {https://arxiv.org/abs/2312.17235} {A simple llm framework for long-range video question-answering}.
\newblock \emph{Preprint}, arXiv:2312.17235.

\bibitem[{Zhang et~al.(2023{\natexlab{b}})Zhang, Han, Liu, Gao, Zhou, Hu, Yan, Lu, Li, and Qiao}]{zhang2023llamaadapter}
Renrui Zhang, Jiaming Han, Chris Liu, Peng Gao, Aojun Zhou, Xiangfei Hu, Shilin Yan, Pan Lu, Hongsheng Li, and Yu~Qiao. 2023{\natexlab{b}}.
\newblock \href {https://arxiv.org/abs/2303.16199} {Llama-adapter: Efficient fine-tuning of language models with zero-init attention}.
\newblock \emph{Preprint}, arXiv:2303.16199.

\bibitem[{Zhong et~al.(2022)Zhong, Ji, Xiao, Li, Deng, and Chua}]{zhong-etal-2022-video}
Yaoyao Zhong, Wei Ji, Junbin Xiao, Yicong Li, Weihong Deng, and Tat-Seng Chua. 2022.
\newblock \href {https://doi.org/10.18653/v1/2022.emnlp-main.432} {Video question answering: Datasets, algorithms and challenges}.
\newblock In \emph{Proceedings of the 2022 Conference on Empirical Methods in Natural Language Processing}, pages 6439--6455, Abu Dhabi, United Arab Emirates. Association for Computational Linguistics.

\bibitem[{Zhu et~al.(2023)Zhu, Chen, Shen, Li, and Elhoseiny}]{zhu2023minigpt}
Deyao Zhu, Jun Chen, Xiaoqian Shen, Xiang Li, and Mohamed Elhoseiny. 2023.
\newblock Minigpt-4: Enhancing vision-language understanding with advanced large language models.
\newblock \emph{arXiv preprint arXiv:2304.10592}.

\end{thebibliography}

\appendix

\section{Licences}
\label{sec:appendix}

We use standard licenses from the community for the datasets, codes, and models that we used in this paper:

\begin{itemize}
    \item \textbf{NExT-QA} \citep{xiao2021next}: \href{https://github.com/doc-doc/NExT-QA/blob/main/LICENSE}{MIT}
    \item \textbf{STAR} \citep{wu2021star}: \href{https://github.com/csbobby/STAR/blob/main/LICENSE}{Apache}
    \item \textbf{How2QA} \citep{li2020hero}: \href{https://github.com/VALUE-Leaderboard/StarterCode/blob/main/LICENSE}{MIT}
    \item \textbf{TVQA} \citep{lei2018tvqa}: \href{https://github.com/jayleicn/TVQA/blob/master/LICENSE}{MIT}
    \item \textbf{IntentQA} \citep{Li_2023_ICCV}: N/A
    \item \textbf{SeViLa} \citep{yu2023self}: \href{https://github.com/Yui010206/SeViLA/tree/main?tab=BSD-3-Clause-1-ov-file}{BSD 3 - Clause}
    \item \textbf{LAVIS} \citep{li2022lavis}: \href{https://github.com/salesforce/LAVIS/blob/main/LICENSE.txt}{BSD 3-Clause}
    \item \textbf{Pytorch} \citep{paszke2019pytorch}: \href{https://github.com/pytorch/pytorch/blob/main/LICENSE}{BSD Style}
    \item \textbf{Q-ViD} (Ours): \href{https://github.com/Daromog/Q-ViD/blob/main/LICENSE}{BSD 3-Clause}

\end{itemize}

\section{Use of Artifacts}
In this work we adopt a open multimodal model, InstructBLIP \citep{dai2023instructblip}, its application in our approach is consistent with its original intended use. For Q-ViD we release our code and we hope it will be useful for future works.

\end{document}